\documentclass[letterpaper]{article} 
\usepackage[submission]{aaai25}  
\usepackage{times}  
\usepackage{helvet}  
\usepackage{courier}  
\usepackage[hyphens]{url}  
\usepackage{graphicx} 
\usepackage{natbib}  
\usepackage{caption} 
\usepackage{algorithm}
\usepackage{algorithmic}

\usepackage{latexsym}
\usepackage{subcaption}
\usepackage{amsmath}
\usepackage{pifont}
\usepackage{multirow}
\usepackage{amsfonts}
\usepackage{makecell}
\usepackage{diagbox}
\usepackage{booktabs}
\usepackage{amssymb}
\usepackage{threeparttable}
\usepackage{newfloat}
\usepackage{listings}
\DeclareCaptionStyle{ruled}{labelfont=normalfont,labelsep=colon,strut=off} 
\floatstyle{ruled}
\newfloat{listing}{tb}{lst}{}
\floatname{listing}{Listing}
\title{An analysis of HOI: \\using a training-free method with multimodal visual foundation models when only the test set is available, without the training set}
\author{
    Written by AAAI Press Staff\textsuperscript{\rm 1}\thanks{With help from the AAAI Publications Committee.}\\
    AAAI Style Contributions by Pater Patel Schneider,
    Sunil Issar,\\
    J. Scott Penberthy,
    George Ferguson,
    Hans Guesgen,
    Francisco Cruz\equalcontrib,
    Marc Pujol-Gonzalez\equalcontrib
}
\affiliations{
    \textsuperscript{\rm 1}Association for the Advancement of Artificial Intelligence\\

    1101 Pennsylvania Ave, NW Suite 300\\
    Washington, DC 20004 USA\\
    proceedings-questions@aaai.org
}

\usepackage{bibentry}

\begin{document}

\maketitle

\begin{abstract}
Human-Object Interaction (HOI) aims to identify the pairs of humans and objects in images and to recognize their relationships, ultimately forming $\langle human, object, verb \rangle$ triplets. Under default settings, HOI performance is nearly saturated, with many studies focusing on long-tail distribution and zero-shot/few-shot scenarios. Let us consider an intriguing problem:``What if there is only test dataset without training dataset, using multimodal visual foundation model in a training-free manner? '' This study uses two experimental settings: grounding truth and random arbitrary combinations. We get some interesting conclusion and find that the open vocabulary capabilities of the multimodal visual foundation model are not yet fully realized. Additionally, replacing the feature extraction with grounding DINO further confirms these findings.
\end{abstract}

\section{Introduction}
\label{sec:intro}
The objective of Human-Object Interaction (HOI) detection is to identify the human pairs in images and discern the relationships between them, which is crucial for various downstream tasks, e.g., visual question answering, robotic vision, embodied intelligence, and video analysis\cite{bemelmans2012socially,dee2008close,feichtenhofer2017spatiotemporal,bolme2010visual}. Although the performance of HOI detection has approached saturation, there remains room for improvement in areas such as zero-shot/few-shot learning, and long-tail distribution. Previous work has achieved promising results by leveraging multimodal visual foundation models for the initialization of textual labels, feature distillation, and so on\cite{gen,eoid}.

Let us consider an interesting question: What results can be achieved using a training-free method with multimodal visual foundation models when only the test dataset is available, without training dataset? We implemented three distinct experimental setups. The first setup involves inputting $\langle human, object \rangle$ pairs from ground truth into a multimodal visual foundation model and comparing these with text prompts containing various verbs to derive a probability distribution of verbs, thus determining the verb outcomes. The second setup disrupts the paired features of ground truth; here, ``human'' comprises all ground truth bounding boxes for humans, and ``object'' includes all bounding boxes. These are then inputted in combinations into a subsequent query module to generate verb results like the first setup. The third setup employs bounding boxes extracted by grounding DINO\cite{liu2023grounding}, which are unpaired, using a method similar to the second setup to ascertain the verb outcomes.

When only test dataset is available and no training dataset exists, all samples are treated equally. In this context, distinctions such as rare/non-rare, seen/unseen combination/object/verb, which become irrelevant.

\begin{figure}[tb]
\centering
\includegraphics[width=0.45\textwidth]{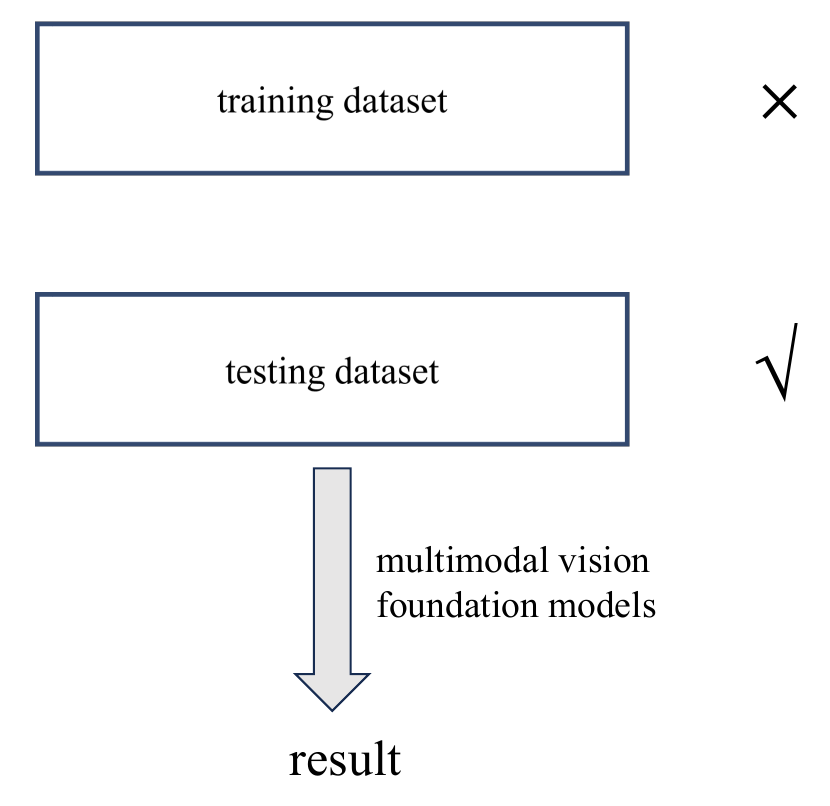}
\caption{Our motivation.}
\label{fig:motivation}
\end{figure}

\begin{figure*}[tb]
\centering
\includegraphics[width=0.95\textwidth]{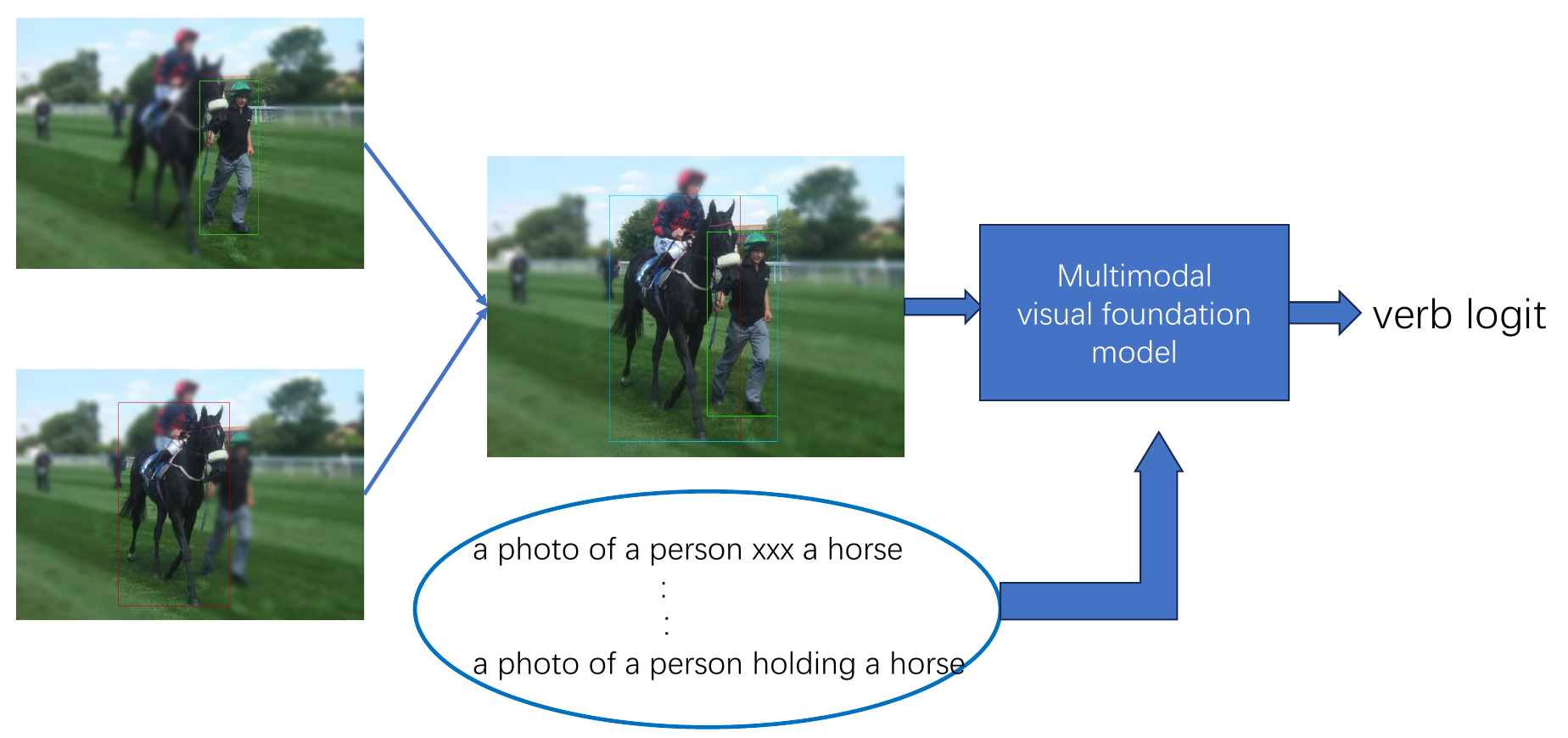}
\caption{The model using the paired ground truth.}
\label{fig:the paired ground truth}
\end{figure*}

In the default experimental setting, rare classes are insensitive to random combinations of humans and objects, whereas non-rare classes demonstrate sensitivity. Under the RF-UC (Rare First Unseen Combinations) setting, tail HOI (rare classes) categories are designated as unseen classes. These unseen classes(rare classes) are insensitive to random combinations, while seen classes(non-rare) are sensitive to random combinations. Conversely, the NF-UC (Non-rare First Unseen Combinations) setting identifies head HOI categories(non-rare) as unseen classes, where unseen classes(non-rare) are sensitive to random combinations, and seen classes(rare) are not sensitive. In experiments involving unseen/seen objects or verbs, the sensitivity to random combinations remains consistent between unseen and seen classifications.

We can conclude that both rare and non-rare classes naturally exist in the environment, each possessing unique properties.
Additionally, our experiments demonstrate that the existing models have not fully developed the zero-shot/few-shot capabilities of multimodal vision foundation models\cite{clip,li2022blip,li2023blip2}.

\section{Related Work}
\label{sec: related_work}
\subsection{HOI}
HOI (Human-Object Interaction) detection\cite{qpic, gen, CDN,hotr,XinpengLiu2022InteractivenessFI} primarily consists of two subtasks: detecting human-object pairs (including their locations and types) and categorizing the types of human-object interactions. HOI detection methodologies are generally divided into two categories: two-stage and one-stage approaches.
In the two-stage approach\cite{IDN,VSG_net,zhong2020polysemy,li2020detailed,kim2020detecting,wan2023weakly}, a separate detector is employed to identify the locations and classes of objects. This is followed by specially designed modules that handle the association of humans and objects and the recognition of their interactions. Conversely, the one-stage approach\cite{hotr,qpic,CDN,XinpengLiu2022InteractivenessFI,gen,eoid,zhong2022towards,wu2022mining,yuanrlip,PenghaoZhou2019RelationPN} involves directly detecting human-object pairs along with their interactions, thereby identifying the corresponding HOI categories in a single step. This paradigm eliminates the need for complex post-processing for human-object matching, enabling end-to-end training.

\subsection{HOI Detection with Linguistic Guidance / zero-shot / few-shot}
Recent advancements in Vision-Language Models (VLMs) have exhibited a promising ability to transfer to downstream tasks\cite{clip,li2022blip,li2022blip2}. The visual representations derived from natural language supervision facilitate zero-shot and open vocabulary tasks. A practical approach to achieve high performance without substantial effort is to utilize information from pre-trained models. An effective strategy involves leveraging linguistic guidance.
With the emergence and demonstrated powerful performance of large-scale pre-trained visual-linguistic models, methods that harness linguistic guidance have shown significant potential in interaction recognition tasks. These tasks necessitate a profound understanding of image context and relational inference. A common technique to incorporate linguistic guidance is to initialize interaction classifiers using text embeddings generated by pre-trained visual-linguistic models\cite{gen,eoid}. Additionally, some studies extract information through knowledge distillation techniques. Directly using predictions from pre-trained visual-linguistic models as constraints is also a favored approach.

\begin{table*}[tb]
    \setlength{\tabcolsep}{1.5pt}
    \centering
        \resizebox{0.97\linewidth}{!}{
        \begin{tabular}{c|ccc|ccc|ccc}
            \hline
            \rule{0pt}{12pt}
            \multirow{2}*{ } &
            \multicolumn{3}{c|}{GT} & \multicolumn{3}{c|}{GT-R} & \multicolumn{3}{c}{GroundingDINO} \\
            \cline{2-10}
            \rule{0pt}{12pt}
            &full	&rare	&non-rare	&full	&rare	&non-rare	&full	&rare	&non-rare \\
            \hline
            CLIP RN101 & 38.14  & 41.02  & 37.28  & 29.75  & 38.00  & 27.28  & 13.56  & 17.99  & 12.23  \\ \hline
        CLIP ViT-B/16 & 41.71  & 46.59  & 40.25  & 32.72  & 43.98  & 29.36  & 15.24  & 21.50  & 13.37  \\ \hline
        CLIP ViT-L/14 & 43.92  & 53.75  & 40.98  & 34.77  & 49.93  & 30.25  & 16.44  & 25.03  & 13.88  \\ \hline
        CLIP ViT-L/14@336px & 44.44  & 54.84  & 41.34  & 35.64  & 51.73  & 30.84  & 17.52  & 28.02  & 14.38  \\ \hline
        blip\_vitB/16 & 40.85  & 44.96  & 39.63  & 31.88  & 41.89  & 28.89  & 15.56  & 19.39  & 14.42  \\ \hline
        blip2\_pretrain\_vitL/14 & 42.79  & 46.96  & 41.54  & 33.23  & 44.77  & 29.78  & 15.34  & 19.39  & 14.12  \\ \hline
        blip2\_pretrain\_vitH/14 & 45.92  & 51.65  & 44.21  & 35.80  & 48.53  & 32.00  & 17.88  & 22.98  & 16.36  \\ \hline
        blip2\_coco\_vitH/14@364px & 49.56  & 54.98  & 47.94  & 38.86  & 50.58  & 35.36  & 19.71  & 24.07  & 18.41 \\
            \hline
        \end{tabular} }
    \caption{mAP on HICO-DET, where``GT'' denotes the input consisting of the ground truth data. ``GT-R'' refers to the input comprising any arbitrary combination of ground truth. ``GroundingDINO'' indicates that the output bounding boxes from the GroundingDINO model are used in any arbitrary combination.}
    \label{mAP on HICO-DET}
\end{table*}

\begin{table}[tb]
    \setlength{\tabcolsep}{1.5pt}
    \centering
        \resizebox{0.97\linewidth}{!}{
        \begin{tabular}{c|ccc|ccc}
            \hline
            \rule{0pt}{12pt}
            \multirow{2}*{ } &
            \multicolumn{3}{c|}{default} & \multicolumn{3}{c}{unseen\_combination} \\
            \cline{2-7}
            \rule{0pt}{12pt}
            &full	&rare	&non-rare	&full	&unseen	&seen \\
            \hline
        GT & 49.56 & 54.98 & 47.94 & 49.56  & 52.45  & 48.84 \\ \hline
        GT-R & 38.86 & 50.58 & 35.36 & 38.86  & 42.29  & 38.00 \\ \hline
        G & 19.71 & 24.07 & 18.41 & 19.71  & 20.33  & 19.56 \\
           \hline
        \end{tabular} }
    \caption{the result1 blip2\_coco\_vitH/14@364px}
    \label{the result1}
\end{table}

\begin{table}[tb]
    \setlength{\tabcolsep}{1.5pt}
    \centering
        \resizebox{0.97\linewidth}{!}{
        \begin{tabular}{c|ccc|ccc}
            \hline
            \rule{0pt}{12pt}
            \multirow{2}*{ } &
            \multicolumn{3}{c|}{rare\_first} & \multicolumn{3}{c}{non\_rare\_first} \\
            \cline{2-7}
            \rule{0pt}{12pt}
            &full	&unseen	&seen	&full	&unseen	&seen \\
            \hline
        GT & 49.56  & 54.66  & 48.29  & 49.56  & 55.23  & 48.14  \\ \hline
        GT-R & 38.86 & 50.49 & 35.95 & 38.86 & 35.11 & 39.8 \\ \hline
        G & 19.71 & 23.96 & 18.65 & 19.71 & 19.97 & 19.65 \\
           \hline
        \end{tabular} }
    \caption{the result2 blip2\_coco\_vitH/14@364px}
    \label{the result2}
\end{table}

\begin{table}[tb]
    \setlength{\tabcolsep}{1.5pt}
    \centering
        \resizebox{0.97\linewidth}{!}{
        \begin{tabular}{c|ccc|ccc}
            \hline
            \rule{0pt}{12pt}
            \multirow{2}*{ } &
            \multicolumn{3}{c|}{unseen\_object} & \multicolumn{3}{c}{unseen\_verb} \\
            \cline{2-7}
            \rule{0pt}{12pt}
            &full	&unseen	&seen	&full	&unseen	&seen \\
            \hline
        GT & 49.56 & 48.29 & 49.82 & 49.56 & 51.26 & 49.29 \\ \hline
        GT-R & 38.86 & 37.03 & 39.23 & 38.86 & 39.36 & 38.78 \\ \hline
        G & 19.71 & 22.71 & 19.12 & 19.71 & 21.96 & 19.35 \\
           \hline
        \end{tabular} }
    \caption{the result3 blip2\_coco\_vitH/14@364px}
    \label{the result3}
\end{table}

\section{Model}
\label{sec: model}
Our study focuses on model performance evaluation in the absence of a training dataset and only test dataset existing, focusing specifically on zero-shot/few-shot, and long-tail distribution recognition capabilities. We operate under the assumption that the feature extraction component of our detection model is predefined. The extracted features are subsequently input into a multimodal visual foundation model to discern relationships between humans and objects. The feature extraction is categorized into three distinct types: features of the paired ground truth, features generated by arbitrarily recombining ground truth pairs, and features extracted using the grounding DINO method, which is not paired.

\subsection{the paired ground truth}
\label{subsec: the paired ground truth}
The model using the paired ground truth is depicted in Figure ~\ref{fig:the paired ground truth}.
In an example, a human is outlined by a green bounding box while a horse is enclosed within a red bounding box. They are identified as a pair, and their union area is subsequently encompassed by a larger blue bounding box. This union area is then inputted into the visual component of a multimodal visual foundational model. Concurrently, the textual description of the image is fed into the textual component of the same model. Ultimately, the output is processed through a softmax function, which yields a probability distribution for different verbs.

\subsection{the arbitrarily recombining ground truth pairs}
\label{subsec: the arbitrarily recombining ground truth pairs}
In this subsection, we did not employ paired ground truth. Instead, we used randomly composed pairs. Specifically, in the $\langle human, object, verb \rangle$ triplet, ``human'' refers to all possible bounding boxes of humans from the ground truth, and 'object' pertains to all possible bounding boxes from the ground truth. The remaining aspects of the model are consistent with those discussed in the previous subsection ~\ref{subsec: the paired ground truth}.

\subsection{grounding DINO}
Using features extracted by grounding DINO, which are not paired, we adopt the same approach as described as ~\ref{subsec: the arbitrarily recombining ground truth pairs}. Here, 'human' refers to all potential bounding boxes of humans as defined by the ground truth, and 'object' relates to all potential bounding boxes as determined by the ground truth.

\section{Experiment}
\label{sec: experiment}
\subsection{datasets}
Our research utilizes the publicly available HICO-DET dataset\cite{hico_det}. HICO-Det comprises 47,776 images, divided into 38,118 for training and 9,658 for testing purposes. It annotates 600 categories of human-object interactions (HOIs), derived from 80 object categories and 117 action categories. Of these, 138 HOI categories, characterized by fewer than 10 training samples each, are classified as Rare. The remaining 462 categories are classified as Non-Rare.

\subsection{results}
The results can be seen in Table~\ref{mAP on HICO-DET}.
The the result in $blip2\_coco\_vitH/14@364px$ can be seen in Table~\ref{the result1}, Table~\ref{the result2}, and  Table~\ref{the result3}.

\subsection{analysis}
When only the test dataset is available, without a corresponding training dataset, all samples are treated uniformly. In this scenario, distinctions such as rare versus non-rare or seen versus unseen combinations of objects and verbs become irrelevant.

In the default setting, rare classes display insensitivity to arbitrary human-object interactions, while non-rare classes are sensitive to these combinations. Under the Rare First Unseen Combinations (RF-UC) setting, rare Human-Object Interaction (HOI) categories are classified as unseen. These unseen (rare) classes exhibit insensitivity to arbitrary combinations, whereas the seen (non-rare) classes remain sensitive. Conversely, the Non-rare First Unseen Combinations (NF-UC) setting assigns non-rare HOI categories as unseen, where these unseen (non-rare) classes are sensitive to arbitrary combinations, while the seen (rare) classes show no sensitivity. Experiments that differentiate between unseen and seen objects or verbs maintain a consistent pattern of sensitivity across these classifications.

\section{Conclusion}
\label{sec: conclusion}
We can deduce that both rare and non-rare classes are inherently present in the environment, each endowed with distinctive characteristics. Moreover, our experiments indicate that current models have not yet fully realized the potential of zero-shot and few-shot learning in multimodal vision foundation models.

{\small
\bibliographystyle{ieee_fullname}
\bibliography{hoibib}
}

\end{document}